\documentclass[a4paper]{jpconf}
\usepackage{graphicx}
\usepackage{wrapfig,lipsum}
\usepackage{amssymb}
\usepackage{amsmath}
\usepackage{color}
\usepackage{hyperref}

\newcommand{\be}{\begin{equation}}

\newcommand{\ee}{\end{equation}} 

\definecolor{bluc}{cmyk}{1,1,0,0.1}
\definecolor{rossoCP3}{cmyk}{0,.88,.77,.40}
\definecolor{rosso}{cmyk}{0,1,1,0.4}
\definecolor{giallo}{cmyk}{0,.33,1,0}
\definecolor{rossos}{cmyk}{0,1,1,0.55}
\definecolor{rossoc}{cmyk}{0,1,1,0.2}
\definecolor{verdes}{cmyk}{0.92,0,0.59,0.4}

\hypersetup{colorlinks, bookmarksopen, bookmarksnumbered,
citecolor=verdes, linkcolor=bluc, pdfstartview=FitH, urlcolor=rossos}

\begin{document}
\title{Testing the boundaries: Normalizing Flows for higher dimensional data sets}

\author{\underline{Humberto Reyes-Gonz\'{a}lez}$^{a,b}$, Riccardo Torre$^{b}$}

\address{$^{a}$Department of Physics, University of Genova, Via Dodecaneso 33, 16146 Genova, Italy \\
$^{b}$INFN, Sezione di Genova, Via Dodenasco 33, I-16146 Genova, Italy}

\ead{humbertoalonso.reyesgonzlez@edu.unige.it, riccardo.torre@ge.infn.it}

\begin{abstract}
Normalizing Flows (NFs) are emerging as a powerful class of generative models, as they not only allow for efficient sampling, but also deliver, by construction, density estimation. They are of great potential usage in High Energy Physics (HEP), where complex high dimensional data and probability distributions are everyday’s meal. However, in order to fully leverage the potential of NFs it is crucial to explore their robustness as data dimensionality increases. Thus, in this contribution, we discuss the performances of some of the most popular types of NFs on the market, on some toy data sets with increasing number of dimensions.
\end{abstract}

\section{Introduction}
In recent years, a variety of generative models have received an exponentially increasing attention by the High Energy Physics (HEP) community, thanks to their ability to encode complex underlying distributions of measured data. Among those, Normalizing Flows (NFs) appear as a particularly interesting brand of generative architectures, as they actually provide an analytic description of the underlying distribution, allowing not only an efficient sampling, but also a destiny estimation of the data. The potential applications in HEP are numerous, since complex probability density functions (pdfs) are found everywhere in the field. However, these pdfs are often high-dimensional complicated functions. Thus, before we start leveraging on the properties of NFs, we wish to answer the question:\textit{ how well do NFs perform in the high-dimensional limit expected from HEP data?}
Therefore, the aim of this contribution is to provide an early answer to this question\footnote{The publication of a full study is left for the near future.}, by testing the currently most widely used NF architectures against complicated toy distributions with increasing dimensions.

%We implemented three of the most widely used NFs architectures and compared their performance against two types of pdfs, namely uncorrelated mixture of gaussians and correlated gaussians, each with 4 to 100 dimensions. 

\section{Normalizing Flows}
Normalizing Flows \cite{9089305} are made of series of bijective, continuous, and invertible transformations that map a simple \textit{base} pdf to a more complicated, \textit{target} pdf. The usual purpose of NFs is to perform a set of transformations, starting from the base pdf, to obtain an approximate description of the unknown underlying distribution of some data of interest. Since the parameters of both the base distribution and the transformations are known, we are able to $generate$ samples of the target distribution, by drawing random points from the base distribution and mapping them into  the target one. This is known as the \textit{generative direction} of the flow. On the other hand, the invertibility of the NF transformations allows to map samples from the target pdf to the base pdf. This is known as the \textit{normalizing direction} of the flow and allows to estimate the probability density of measured data described by the target distribution.

The basic principle of the NFs is just the formula for the change of variables in a pdf. Let $Z \in \mathbb{R}^{D}$ be a random variable with a known tractable pdf $p_Z:  \mathbb{R}^{D} \rightarrow \mathbb{R}^{D}$.  Suppose we want to map $p_Z$ to the pdf $p_Y$ of a different random variable $Y\in \mathbb{R}^{D}$. For this, we only need to find a bijective function $\mathbf{g}$, with inverse $\mathbf{f}$, which satisfies $Y=\mathbf{g}(Z)$. Then, $p_{Z}$ is mapped to $p_{Y}$ through the relation
\be \label{NF1}
p_Y(y)=p_Z(\mathbf{f}(y))|\det \mathrm{J}_{f}|=p_Z(\mathbf{f}(y))|\det \mathrm{J}_{g}|^{-1}\,,
\ee
where  $\mathrm{J}_{f}=\frac{\partial \mathbf{f}}{\partial y}$ is the Jacobian of $\mathbf{f}$ and $\mathrm{J}_{g}=\frac{\partial \mathbf{g}}{\partial z}$ is the Jacobian of $\mathbf{g}$. Furthermore, $\mathbf{g}$ can be generalized to a set of $N$ transformations as $g=g_N\circ g_{N-1}\circ ...g_1$ with inverse $f=f_1\circ ... f_{N-1} \circ f_N$ and $\det \mathrm{J}_{f}=\prod_{i=1}^{N}\det \mathrm{J} _{f_i}$, where each $\mathbf{f}_i$ depends on a  $y_i$ intermediate random variable. 

Given that $\mathbf{g}$ is an invertible function with known parameters $\theta$, and $p_Z$ is a simple distribution described by some parameters $\phi$, we can perform likelihood-based density estimation on the measured data set $\mathcal{D}=\lbrace y^{(i)}\rbrace^{M}_{i=1}$. For a single transformation flow, the log-likelihood of the data is simply
\be
\mathrm{log} p(\mathcal{D}|\Theta)=\sum_{i=1}^{M}\mathrm{log} p_{Y}(y^{(i)}|\Theta)=\sum^{M}_{i=1}\mathrm{log}p_{Z}(\mathbf{f}(y^{(i)}|\theta)|\phi)+\mathrm{log}|\mathrm{det}J_{f}|.
\ee
The remaining matter is then to find a set of parameters $\Theta=\lbrace\phi,\theta\rbrace$ that can accurately describe the underlying probability distribution of the measured data. The choice of parameters must satisfy the following three conditions:
\begin{itemize}
\item they must be invertible;
\item they must be sufficiently expressive to model the target distribution;
\item they should be computationally efficient, i.e.~the transformations should be chosen such that $\mathbf{g}$, $\mathbf{f}$, and the determinant of the Jacobians are easily computed.
\end{itemize}

\section{Coupling and autoregressive flows}
There is a wide and growing variety of NF architectures in the market (see ref.~\cite{9089305} for an overview), which satisfy the conditions listed above. In most, if not all of them, the parameters $\Theta$ are learned by neural networks\footnote{In practice, is often the case that $\phi$ is fixed and only $\theta$ is modeled by the neural network.} via minimization of the loss function $-\mathrm{log} p(\mathcal{D}|\Theta)$. In this contribution, we focus on Coupling and Autoregressive Flows; currently the most widely used implementations of NFs.
\paragraph{Coupling flows.} Consider a disjoint partition of a random variable $Y\in  \mathbb{R}^{D} $ such that $(y^A,y^B)\in  \mathbb{R}^{d}\times  \mathbb{R}^{D-d}$ and a bijection $\mathbf{h}(\cdot;\theta):  \mathbb{R}^{d}\rightarrow  \mathbb{R}^{d}$. Then we can define $\mathbf{g}$ such that
\be
\begin{array}{l}
y^{B}=z^B\,,\vspace{1mm} \ \ \
y^{A}=h(z^A;\Theta(z^B))\,,
\end{array}
\ee
where the parameters $\theta$ are defined by the \textit{conditioner} $\Theta(z^B)$,  usually modeled by a neural network that uses $z^B$ as input. The \textit{coupling function} $\mathbf{h}$ must be invertible such that the \textit{coupling flow} $\mathbf{g}$ also fulfills the condition. The inverse of $\mathbf{h}$ then follows
\be
\begin{array}{l}
z^{B}=y^B\,,\vspace{1mm} \ \ \
z^{A}=h^{-1}(y^A;\Theta(z^B))\,.
\end{array}
\ee
The jacobian of $\mathbf{g}$ is then a two block triangular matrix which, for dimensions $\lbrace 1:d \rbrace$, is given by the Jacobian of $\mathbf{h}$  $\mathrm{J}_h$, and for dimensions $\lbrace d:D\rbrace$ is an identity matrix. Thus, $J_g$ is simply  $ \mathrm{det} J_g=\Pi_{i}^{d} \frac{\partial h_i}{\partial z^{A}_i}$.

%\be
 %\mathrm{det} J_g=\mathrm{det} J_h=\Pi_{i}^{d} \frac{\partial h_i}{\partial z^{A}_i}.
%\ee

\paragraph{Autoregressive flows.} Consider now a bijector $\mathbf{h}(\cdot;\theta):  \mathbb{R}\rightarrow  \mathbb{R}$, again parametrized by $\theta$. We can define an \textit{autoregressive flow} $\mathbf{g}$ such that
\be\label{eq:autoreg1}
\begin{array}{l}
y_1=z_1\,,\vspace{1mm} \ \ \
y_j=h(z_j;\Theta_{j}(z_{1:j-1}))\,,
\end{array}
\ee
where $j=\lbrace 2,3,...,D\rbrace$. The parameters of the conditioners $\Theta_{j}$ are thus modeled by neural networks that take the previous dimensions of $Z$ as input. The resulting Jacobian of $\mathbf{g}$ is again a triangular matrix, now given by $\mathrm{det} J_g=\Pi_{j=2}^{D}\frac{\partial h_j}{\partial z_j}\,$. The inverse of $\mathbf{h}$ yields
\be
\begin{array}{l}
z_1=y_1\,,\vspace{1mm}  \ \ \
z_j=h^{-1}(y_j;\Theta(z^{1:j-1}))\,.
\end{array}
\ee 
%\be
%\mathrm{det} J_g=\Pi_{j=2}^{D}\frac{\partial h_j}{\partial z_j}\,.
%\ee

 Finally, note that $\Theta_{j}$ can also be determined with the \textit{previous} dimensions of $Y$, such that $y_j=h(z_j;\Theta(y_{1:j-1}))$. The choice of variables used to model the conditioner will depend on whether the NF is intended for sampling or density estimation. In the former case, $\Theta$ is usually chosen to be modeled by the base variable $Z$. In this way, the transformations in the generative direction would only require one forward pass through the flow, while transformations in the normalizing direction would require $D$ iterations trough the autoregressive architecture. Inversely, for densitiy estimation, is often convenient to parametrise the conditioner using the target variable $Y$, as now the normalizing transformations are direct as opposed to the resulting autoregressive generative transformations. 
 
 \section{The flows under scope}
For this study, we have chosen three of the most popular implementations of coupling and autoregressive flows in the market. Namely, the Real-valued Non-Volume Preserving (RealNVP) \cite{dinh2017density}, the Masked Autoregressive Flow (MAF) \cite{papamakarios2018masked}, and the Autoregressive-Rational Quadratic Spline (A-RQS) \cite{durkan2019neural}.
 
 The RealNVP are a type of coupling flows whose bijectors $\mathbf{h}$ are affine functions such that
\be
\begin{array}{l}
y_{1:d}=z_{1:d}\,,\vspace{1mm} \ \ \
y_{d+1:D}=z_{d+1:D}\odot\exp (s(z_{1:d}))+t(z_{1:d})\,,
\end{array}
\ee
where $s$ and $t$ respectively correspond to the scale and translation transformations modeled by the neural network. The determinant of the Jacobian is simply $\det J =\Pi_{i=1}^{D-d}s_{i}(z_{i:d})$  and the inverse of $\mathbf{h}$ yields $z_{d+1:D}=(y_{d+1:D}-t(y_{1:d})/\exp (s(y_{1:d}))\,.$

The choice of the partition is not uniquely defined. In principle, the size of $d$ may vary between $1$ and $D-1$. Furthermore, the partition criteria can follow different patterns. For instance, in ref.~\cite{dinh2017density} a binary mask is implemented as partitioner, which is either defined as a spatial checkerboard pattern or a channel-wise masking.

The MAF are autoregressive flows where the bijectors are also affine functions. The flows are described as
\be\label{eq:MAF1}
\begin{array}{l}
y_{1}=z_{1}\,,\vspace{1mm} \ \ \
y_{j}=z_{j}\odot\exp (s(y_{1:j-1}))+t(y_{1:j-1})\,.
\end{array}
\ee
The determinant of the Jacobian is simply $\det J =\Pi_{i=1}^{D}s_{i}(z_{i})$ and the inverse of $\mathbf{h}$ yields  $z_{j}=(y_{j}-t(y_{1:j-1})/\exp (s(y_{1:j-1}))\,.$

Note that the MAF bijectors are iteratively  modeled with the \textit{previous} dimensions of $Y$. Thus, making them suitable for density estimation. Fortunately, the MAF architecture efficiently computes all the entries of the flow in a single pass to a neural network featuring a different masking pattern for each dimension; hence, the name \textit{Masked} Autoregressive Flows.

The A-RQS. Coupling and autoregresive flows are not restricted to affine functions. It is possible to implement more expressive bijectors as long as they are invertible and are sufficiently computationally efficient. A very interesting example are the monotonic Rational Quadratic Splines bijectors. The resulting coupling flows are made of $K$ bins, where each bin is defined by a monotonically-increasing rational-quadratic function. The whole spline is defined between an interval $[-B,B]$, outside of which is defined as an identity transformation. The boundaries of each of the bins are set by $K+1$ \textit{knots} with coordinates $\lbrace (x^{(k)},y^{(k)})\rbrace^{K}_{k=0}$. At each internal knot, arbitrary positive values $\lbrace \delta^{(k)}\rbrace_{k=1}^{K-1}$  are given to the corresponding derivatives, while at the boundaries of the spline the derivatives are fixed to $\delta^{(0)}=\delta^{(K)}=1$. By defining $ s_{k} = (y^{k+1} - y^{k})/(x^{k+1} - x^{k})$ and $ \xi(z) = (z - x^{k}) / (x^{k+1} - x^{k}) $, for the $k^{th}$ bin, the corresponding rational-quadratic function $ \alpha^{(k)}(\xi) / \beta^{(k)}(\xi) $ is defined as
\be\label{eq:rq-transformation}
    \frac{\alpha^{(k)}(\xi)}{\beta^{(k)}(\xi)} = y^{(k)} + \frac{(y^{(k + 1)} - y^{(k)}) (s^{(k)} \xi^{2} + \delta^{(k)} \xi (1 - \xi))}{s^{(k)} + (\delta^{(k + 1)} + \delta^{(k)} - 2 s^{(k)}) \xi (1 - \xi)}.
\ee
%The jacobian is given by
%\begin{align}
%	\frac{\partial}{\partial y}(\frac{\alpha^{(k)}(\xi)}{\beta^{(k)}(\xi)} )=  \frac{(s^{(k)})^{2} (\delta^{(k + 1)} \xi^{2} + 2 s^{(k)} \xi (1 - %\xi) + \delta^{(k)} (1 - \xi)^{2} )}{(s^{(k)} +(\delta^{(k + 1)} + \delta^{(k)} - 2 s^{(k)}) \xi (1 - \xi))^{2}}.
%\label{eq:rq-derivative}
%\end{align}
%where the derivative is performed with respect of the random variable $Y$ as usual. Finally, the inverse is given by one of the %quadratic root solutions given by $ \xi(x)=2c/(-b-\sqrt{b^2-4ac})$, where
%inverse is: 
%\begin{align}
 %  a =  (y^{(k + 1)} - y^{(k)}) (s^{(k}) - \delta^{(k)}) +( y - y^{(k)} ) (\delta^{(k + 1)} + \delta^{(k)} - 2 s^{(k)}),\\
%	b = (y^{(k + 1)} - y^{(k)})\delta^{(k)} -  ( y - y^{(k)} )(\delta^{(k + 1)} + \delta^{(k)} - 2 s^{(k)}),\\
%	c = - s^{(k)} (y - y^{(k)} ),
%\end{align}
In practice, $B$ and $K$ are fixed variables, while $\lbrace (x^{(k)},y^{(k)})\rbrace^{K}_{k=0}$ and $\lbrace \delta^{(k)}\rbrace_{k=1}^{K-1}$ are the parameters modeled by the neural network. Furthermore, although we are now left with slightly more complex Jacobians and inverse functions, they can still be efficiently computed.  More details on how this is done, can be found in ref.~\cite{durkan2019neural}. Finally, we stress that RQS can be implemented as coupling and autoregressive flows. However, for this early study, we only focus on the autoregressive version.

\begin{table}
\caption{\label{tab:table} Hyper-parameters that led to the best performances when learning MoG (left) and CG (right)  distributions for each of the NF architectures.}
\parbox[t]{.45\linewidth}{\footnotesize
\centering
\begin{tabular}{llll}

\br
NF &  N bij& Hidden layers & N samples\\
\mr
MAF& 3,5,10  &  $128,256\times 3$  &$100k, 300k$\\
RealNVP &  10 &  $128,256 \times3$  & $100k, 300k$  \\
A-RQS (8knots) & 2 &  $128,256\times 3$  &$100k$\\
\br
\end{tabular}
%\caption{\label{label}Table caption.}
}
\hfill
\parbox[t]{.45\linewidth}{\footnotesize
\centering
\begin{tabular}{cccc}

\br
NF &   N bij& Hidden layers & N samples\\
\mr
MAF& 3,10  &  $32,64,128\times 3$  &$100k, 300k$\\
RealNVP &  3,10 &   $32,64,128\times 3$  & $100k, 300k$  \\
&&&\\
\br
\end{tabular}
}
\vspace{-5mm}
\end{table}
\normalsize

\section{The test setup}
To test the robustness of the NF architectures under scope, we have implemented two types of target pdfs. The first ones are {Uncorrelated Mixture of three Gaussians (MoG), where each of the $D$ dimensions is parametrised by three means and variances. The purpose of the MoG is to test the flows' capabilities of learning multi-modal shapes in high dimensional data sets. The second set of distributions are  Correlated Gaussians (CG), parametrised by $D$ means, one per dimension, and a $D\times D$ correlation matrix. They are intended to test the NFs robustness against randomly correlated data sets. For both types of distributions we have implemented versions with increasing number of dimensions $D=\lbrace 4,8,16,32,64,100 \rbrace$.

The next piece of the testing setup, is the measure under which we test and compare the performances of the Normalizing Flows. Although we are generating data samples from well defined toy distributions, and thus know analytically their pdfs, in real life we don't know the pdf of measured data (this is actually the reason to use NFs). Hence, we require metrics that don't rely on directly comparing analytic pdfs. Instead, we use metrics based on non-parametric tests, as they are agnostic about the parameters of the underlying pdfs. For this reason we have chosen to test the performance of our NFs with the 2-sample 1D Kolgomonov-Smirnov test (KS-test) and the 1D Wasserstein distance metrics, which rely only on empirical distributions functions. The KS-test computes the $p$-value for two 1-dimensional samples to come from the same \textit{unknown} pdf. In practice, we perform the test for several small samples and average over each of the $D$ dimensions. Then, we compute the median KS-test out of the $D$ $p$-values. Since the $p$-value is a random variable with a uniform distribution between zero and one and this is a two-sided measure, the optimal value is $0.5$. The W-distance is usually described as the minimal amount of \textit{work} required  to transform one pdf into another. Again, we take the median across the $D$ dimensions. The optimal value in this case is 0. Additionally, to estimate how well the correlations are learned by the NF, we computed the Frobenius norm (F-norm) of the subtraction between the correlation matrices corresponding to the `real' and `learned' distributions. The obtained values were then standardized by dividing over the $(D^2-D)/2$ number of elements in the matrices.

Regarding the training and hyper-parameter setups, all the NFs were trained for three iterations, each one with 300 epochs and an early stopping callback with patience 30. The starting learning rate was set to .001 with a dropping rate of $1/10$ after each iteration. Moreover, we always used the \textsc{Adam} optimizer and the \textsc{Relu} activation function for the hidden layers. Furthermore, between each NF layer a permutation bijector was implemented to ensure that all dimensions are trained. Finally, for the A-RQS we fixed the number of knots to $K=8$ and the boundaries as $B=12$. 

To optimize each NF architecture we performed a small scan over the remaining main free hyper-parameters. Namely, the number of NF bijectors, the number and size of the hidden layers in each neural network and the  number of training samples. The combinations that generally provided the best results are shown in Table \ref{tab:table}, and are discussed in the next section.

All the results were obtained using \textsc{Tensorflow 2} and \textsc{Tensorflow Probability}. As a reference, the NFs were trained on a  Nvidia Tesla-V100 32Gb GPU.

\section{Results.}

\begin{figure}[t!]
\begin{center}
\includegraphics[scale=.41]{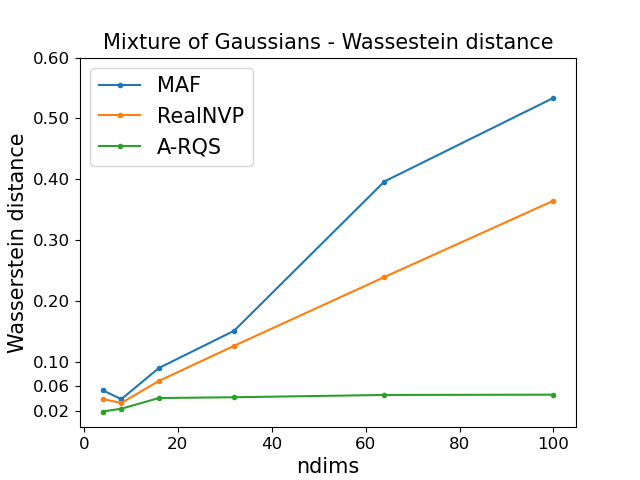} 
\includegraphics[scale=.41]{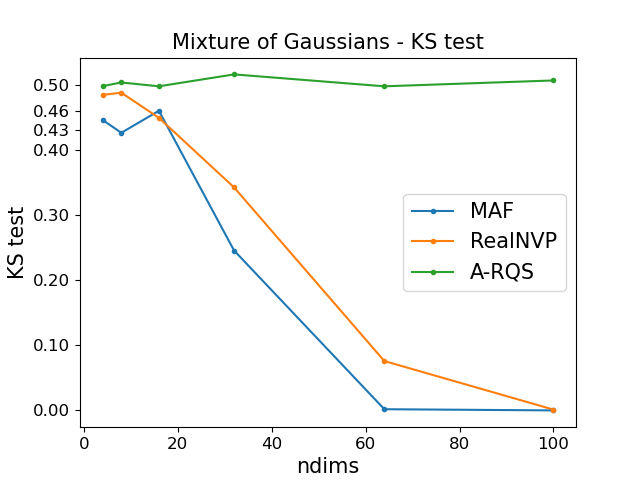} 
\end{center}
\vspace{-5mm}
\caption{Performance comparison between the RealNVP, MAF and A-RQS architectures when learning the MoG distributions, measured by the W-distance (left) and the KS-test (right).}\label{fig:mixgauss}
\vspace{-5mm}
\end{figure}
Performances for the NF modeling of the MoG distributions are shown in Figure \ref{fig:mixgauss}, as measured by the W-distance (left panel) and KS-test (right panel) metrics. Note that we always find consistent results between the two metrics. The best performing NF is by far the A-RQS, which yielded KS-tests $\backsim .5$ and W-distances $\backsim 0$, for all MoGs. This is true even for simple hyper-parameters setups. As shown in Table \ref{tab:table}, already with  only 2 bijectors and $100k$ training samples optimal results are obtained. In almost all cases the $128\times 3$ hidden layer shape was enough, the only exception being actually at $D=4$, where a $256\times 3$ hidden layer shape was required. Moreover, the training time needed by the A-RQS to achieve such results was always of the order of few minutes. Regarding the NF architectures with affine function bijectors, we obtain good results for both of them up to $D=16$. However, we find a linear decrease in training accuracy as the dimensionality increases. This remained true even for hyper-parameter setups with 10 bijectors, $256\times 3$ hidden layers and $300k$ training samples.
%Augmenting the number of samples and the hidden layer complexity above  and , respectively, did not yield significant improvement but substantially enlarged the training time.

Concerning the NF modeling of the CG distributions, since we are now mapping (uncorrelated) gaussians to (correlated) gaussians, we are now only interested in the NF's capability of estimating the correlations within pdf dimensions, not involving multi-modal distributions. 
%We turn to the performance of the NF architectures when modeling the CG distributions. Note that now we are only interested in the NF's capabilities for estimating the correlations within pdf dimensions, thus we are mapping (uncorrelated) gaussians to (correlated) gaussians; not involving non-linear distributions. 
For this reason we only compare the MAF vs the RealNVP architectures, i.e.~coupling vs autoregressive flows. Actually, the complexity of the A-RQS architectures leads to training instabilities when applied to such simple transformations. Results are shown in Figure \ref{fig:corrgauss}, as measured by the W-distance (left panel) and KS-test (center panel) metrics and the F-norm (right panel). Again, an overall consistency between the KS-test and W-distance metrics is found. The results correspond to NFs with up to 10 bijectors, $128\times 3$ hidden layers and $300k$ training samples. Regarding the RealNVP, good performance is obtained for $D\leq8$, while for $D>8$ accuracy starts to noticeably drop, when looking at the non-parametric metrics. As for the F-norm, the worst result was found for $D=16$, with F-norm$\sim .0028$. Turning to the MAF, the non-parametric tests show quite good results for $D\leq16$, while for $D> 16$ the performance starts to decrease but at a much lower rate. Whereas the F-norms are always below $.0018$,  with worst value found at $D=8$. Regarding training times, the autoregressive flows required $\sim$30mins at most, while coupling flows took a long as 140mins to be trained.
%the KS-test/W-distance is always above/below .46/.4.

\begin{figure}[t!]
\begin{center}
\includegraphics[scale=.32]{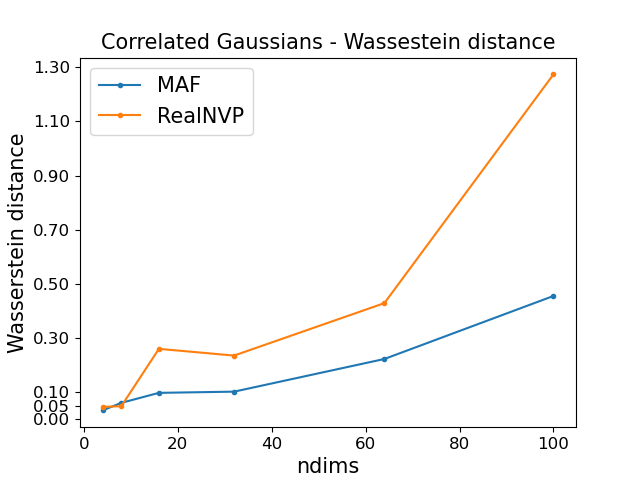} 
\includegraphics[scale=.32]{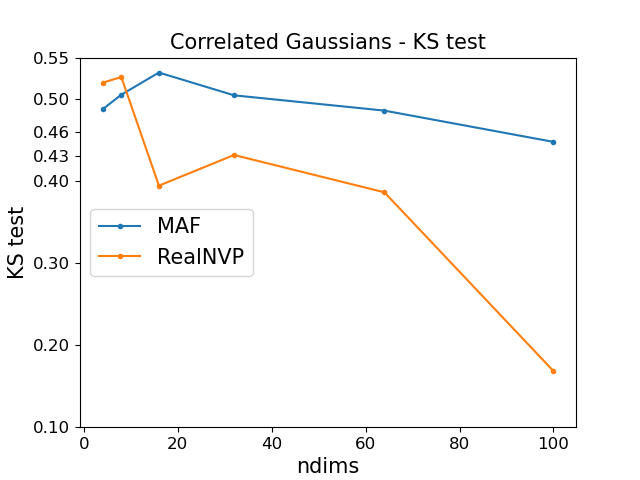} 
\includegraphics[scale=.32]{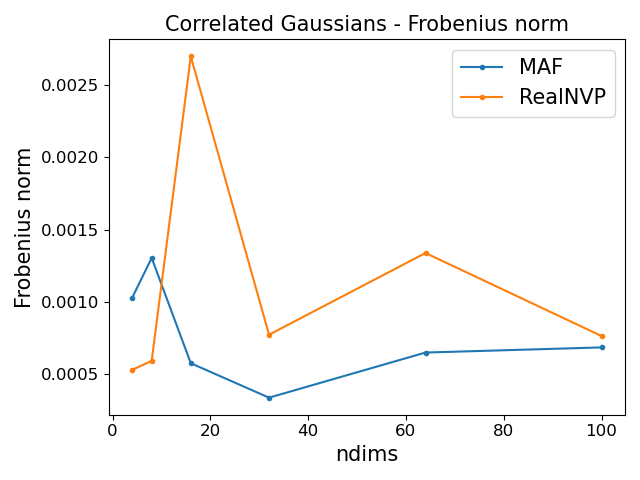}
\end{center}
\vspace{-5mm}
\caption{Performance comparison between the RealNVP and MAF architectures when learning the CG distributions, measured by the W-distance (left), the KS-test (center) and the F-norm (right). }\label{fig:corrgauss}
\vspace{-5mm}
\end{figure}
\section{Conclusion}
Given the interest in using NF architectures to describe complicated high-dimensional pdfs found in HEP,  we investigated, on more general grounds, how well would the current state-of-the-art NFs perform when modeling high dimensional complicated or heavily correlated pdfs. To this aim, we have performed an early study where we tested the most widely used NF architectures, the so-called coupling and autoregressive flows, against two sets of toy distributions with increasing dimensions, one with multi-modal shapes and one with random correlations. As a highlight, we found that the A-RQS architecture is expressive enough to accurately model these high-dimensional non-linear distributions in a very short time. 

%The first set is designed to test the NF's performance when learning non-linear distributions. As a highlight, we found that the A-RQS architecture is expressive enough to accurately model these high-dimensional non-linear distributions in a very short time. The second set is made of randomly correlated normal distributions. With them, we compared the coupling (RealNVP) and autoregressive (MAF) flows' abilities to describe correlated pdfs. We found that, albeit good results are found at low dimensions, accuracy starts to decrease when the dimensionality of the pdf increases. However, the MAF presented a milder performance descent as compared to the RealNVP. Nonetheless, Is our expectation that larger values of the free hyper-parameters should yield an improve performance.

Finally, we stress that these are early results and a more detailed study is being carried out. Among other things, we plan to enlarge the scope of the hyper-parameters and toy distributions, possibly include more architectures, and discuss the NF's potential usage for HEP data augmentation.
%Among other things, we plan to enlarge the scope of the hyper-parameters, include the coupling version of the RQS.

\section*{Acknowledgements}
This research was funded by the Italian PRIN grant 20172LNEEZ.
\section*{References}
%bibliographystyle{plain}
%\bibliography{references}

\end{document}